\pdfoutput=1
\documentclass[letterpaper, 10pt, conference]{ieeeconf}
\IEEEoverridecommandlockouts

\usepackage{cite}
\usepackage{amsmath,amssymb,amsfonts}
\usepackage{graphicx}
\usepackage{float}
\usepackage{booktabs}
\usepackage{multirow}
\usepackage{xcolor}
\usepackage{colortbl}
\usepackage{tabularx}
\usepackage{threeparttable}
\definecolor{oursrow}{gray}{0.92}
\usepackage[T1]{fontenc}
\usepackage{textcomp}
\usepackage[hidelinks,breaklinks=true]{hyperref}
\usepackage{url}

\newcommand{\token}{$\langle$\texttt{gait}, \texttt{speed}, \texttt{social\_cost}$\rangle$}
\newcommand{\acr}[1]{\textbf{#1}}  

\newcommand{\captoppad}{\vspace*{5pt}}
\newcommand{\figtoppad}{\vspace*{4pt}}

\makeatletter
\let\TACET@makecaption\@makecaption
\long\def\@makecaption#1#2{%
  \ifx\@captype\@IEEEtablestring
    {\centering\footnotesize{\scshape #1:}\ #2\par}%
    \@IEEEtablecaptionsepspace
  \else
    \TACET@makecaption{#1}{#2}%
  \fi}
\makeatother

\makeatletter
\def\@IEEEfigurecaptionsepspace{\vskip -1.5pt\relax}
\makeatother

\begin{document}

\title{TACET: Context-Appropriate Acoustic-Social Navigation for Quadrupeds}

\author{Sungsan Park$^{1}$, Young-Sik Shin$^{2,*}$, and Sanghyun Kim$^{1,*}$
\thanks{This research was supported by the National Research Foundation of Korea (NRF) grant funded by the Korea government (MSIT) (No.~\mbox{RS-2024-00411007}, \mbox{RS-2026-25618697}) and by the Korea Basic Science Institute (National Research Facilities and Equipment Center) grant funded by the Ministry of Science and ICT (No.~\mbox{RS-2025-00564593}).}
\thanks{$^{1}$Department of Mechanical Engineering, Kyung Hee University, Yongin, 17104, Republic of Korea. {\tt\small \{san2642, kim87\}@khu.ac.kr}}%
\thanks{$^{2}$School of Mechanical Engineering, Kyungpook National University, Daegu, 41566, Republic of Korea. {\tt\small yshin86@knu.ac.kr}}%
\thanks{$^{*}$Corresponding authors: Young-Sik Shin and Sanghyun Kim.}%
}

\maketitle

\begin{abstract}
Quadruped robots entering hospitals, care homes, and quiet offices must be
context-appropriate not only in \emph{where} they move but in \emph{how loudly}
they move: a legged robot's locomotion noise, dominated by foot--ground impacts,
is itself a social variable. Prior social navigation respects human space but
treats the robot as acoustically uniform, while quiet-locomotion methods reduce
noise to an operator-specified, context-blind level. We present \textbf{TACET}, a context-appropriate acoustic-social navigation method that
infers social context from the robot's egocentric view and decides both where it walks and
how loudly, coupling a slow fine-tuned vision-language reasoner to a fast reactive
controller through a single compact behavior token, \token. The same token conditions both a social costmap (\emph{where} to go) and
a quiet locomotion policy (\emph{how loudly} to move), while a structured
out-of-view memory keeps recently seen people in the reasoner's context after they
leave the camera view. On a real quadruped, context-conditioned locomotion lowers
locomotion noise by up to 9.3\,dBA at matched speed, and across our scenarios the full method
keeps personal-space compliance at 100\% with low acoustic intrusion ($\le$2.9\,dBA),
jointly improving spatial and acoustic performance in the evaluated scenarios.
The project page is available at
\url{https://rcilab.khu.ac.kr/tacet/}.
\end{abstract}

\section{Introduction}
\begin{figure}[t]
\centering
\figtoppad
\includegraphics[width=\columnwidth]{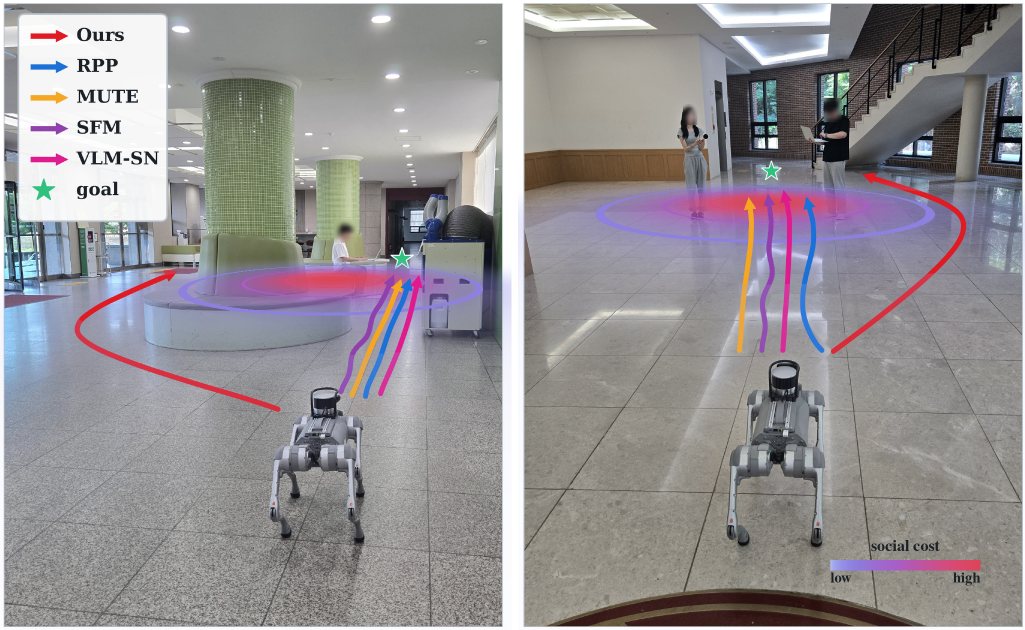}
\caption{Method paths on real runs of the two indoor scenarios: a focused person working
at a table (left) and a standing group in conversation (right). Shading is the social costmap,
rendered on the ground plane. TACET (red)
swings wide and keeps its distance, whereas RPP (blue), MUTE (amber) and VLM-Social-Nav
(magenta) take the short path past the people, and SFM (purple) enters the personal zone
and passes closest to the seated person.}
\label{fig:teaser}
\end{figure}

Consider a quadruped crossing a shared building lobby
(Fig.~\ref{fig:teaser}). A geometry-only planner treats each person as an obstacle
and takes the shortest collision-free path, brushing past someone absorbed in
work at a table, then threading between people standing in conversation, its footfalls
fully audible. The path is safe, yet it is
socially inappropriate in two distinct ways: \emph{where} the robot moves, as
it crowds personal space and cuts through a conversation, and \emph{how loudly} it
moves, as its footfalls intrude on people who need quiet. Unlike wheeled
platforms, a legged robot is intrinsically loud because its locomotion noise is
dominated by foot--ground impacts~\cite{mute}, so its acoustic footprint is a
social variable that must be managed online.

Prior work largely follows two threads, each addressing only half of this problem. Social navigation makes the robot respect human space, but treats it as acoustically uniform~\cite{vlmsocialnav,falcon}; quiet locomotion makes a quadruped quieter, but to an operator-specified degree that is blind to context~\cite{mute}. Acoustic-aware planning relaxes this by predicting how loud the robot sounds to nearby listeners, yet it needs a prior visual scan of the environment and manages noise at the level of the path rather than the robot's own gait~\cite{anavi}. To our knowledge, no prior method has demonstrated social-context reasoning coupled to a controllable acoustic footprint: the former knows where to go but not how quiet to be, while the latter can be quiet but not when to be.

To this end, we propose \textbf{TACET} (\acr{T}oken-aware \acr{A}cousti\acr{C}-social \acr{E}tiquette \acr{T}raversal), a context-appropriate acoustic-social navigation method for quadruped robots that closes this gap. A slow semantic reasoner, a fine-tuned vision-language model (VLM), reads the robot's egocentric view together with a structured memory of nearby people, including those who have just left the camera's field of view, and emits a compact behavior token with three fields: \texttt{gait}, \texttt{speed}, and \texttt{social\_cost}. A fast reactive controller consumes this token to shape both a social costmap, which governs \emph{where} to go, and a token-conditioned quiet locomotion policy, which governs \emph{how loudly} to move, while a standard real-time planning stack preserves safety. The two processes are coupled only through this small token interface, allowing rich but slow social reasoning to steer real-time control without entering the safety loop. Because the same token scales spatial avoidance and acoustic effort together, the robot becomes context-appropriate rather than unconditionally quiet or fast.

\noindent Our main contributions are as follows:
\begin{enumerate}
\item We introduce \emph{context-appropriate acoustic-social navigation}: a robot infers social context and modulates not only \emph{where} but also \emph{how loudly} it moves, unifying spatial and acoustic compliance, previously treated in isolation, under one inferred signal.
\item We realize it as TACET: a fine-tuned VLM emits a compact behavior token that jointly drives a social costmap with out-of-view memory and a token-conditioned quiet locomotion policy, linking slow reasoning to a real-time controller through a small, interpretable interface.
\item We evaluate TACET on a Unitree Go2, where it reduces locomotion noise by up to
9.3\,dBA at matched speed through context-conditioned locomotion, and it is the only method to keep full Personal Space Compliance in every
scenario, at a closest approach of 1.44--3.62\,m against at most 1.35\,m for any
baseline, showing that
acoustic and social compliance can be attained together rather than traded off.
\end{enumerate}

\section{Related Work}
\subsection{Social Robot Navigation}
Classical social navigation builds on geometric planners such as the social-force
model~\cite{sfm}, which are efficient and well-validated but treat people as dynamic
obstacles, yielding socially blunt behavior. On the real-world SCAND dataset, geometric systems already match human demonstrations in
many situations, arguing for augmenting rather than discarding them~\cite{raj}. Our
design inherits this view: we retain a validated real-time planning stack for
reactive safety and add VLM-based semantic reasoning above it.

Learning-based methods instead imitate human navigation or optimize social objectives with
reinforcement learning; Falcon, for example, predicts human trajectories and penalizes
actions that obstruct them~\cite{falcon}. These methods motivate the proxemics-grounded
metrics we adopt, but they operate kinematically in simulation and reason only about
\emph{where} to go, never the gait, foot--ground contact, or acoustics a physical quadruped
must manage.

More recently, VLMs have brought open-vocabulary social reasoning to navigation.
VLM-Social-Nav queries a VLM-based scoring module for a social cost term that corrects a
low-level planner, structuring the VLM output into linguistic heading and speed terms
mapped to linear and angular velocity~\cite{vlmsocialnav}. Narrate2Nav instead embeds language reasoning implicitly in a visual encoder via a Barlow-Twins objective, enabling real-time RGB-only navigation that reasons in latent rather than token space~\cite{narrate2nav}. In contrast to these spatial- and trajectory-level methods, which keep people's space but assume an acoustically uniform robot and largely target wheeled platforms, TACET conditions both a social costmap and a quadruped's quiet gait on VLM-inferred context, making the robot's acoustic footprint itself a function of social context.

\subsection{Quiet and Acoustic-Aware Locomotion for Legged Robots}
The dominant source of quadruped locomotion noise is the foot--ground impact,
whose acoustic amplitude grows with foot velocity at contact~\cite{mute}. MUTE reduces it through control alone, with a quiet factor that trades noise reduction
against agility and an evaluation protocol based on Mean and Peak Noise Levels (MNL, PNL)
at matched actual speed~\cite{mute}. The desired quietness there is an
operator-specified parameter: the robot does not perceive who is nearby, and so cannot
determine \emph{when} quietness is socially required.

On the acoustic-aware navigation side, ANAVI is closest to our motivation: it visually
estimates how loud the robot will sound at a listener's location and uses this as a cost to
plan a quieter route~\cite{anavi}. It nonetheless depends on a prior visual scan of the environment and manages
noise at the level of the path rather than the gait, and the social weighting it proposes
is left unevaluated. In contrast, the token-conditioned policy in TACET is driven by VLM-inferred social context and decides online, without any prior scan of the environment, both when and how much to attenuate its own locomotion noise.

\begin{figure*}[!t]
\centering
\figtoppad
\includegraphics[width=\textwidth]{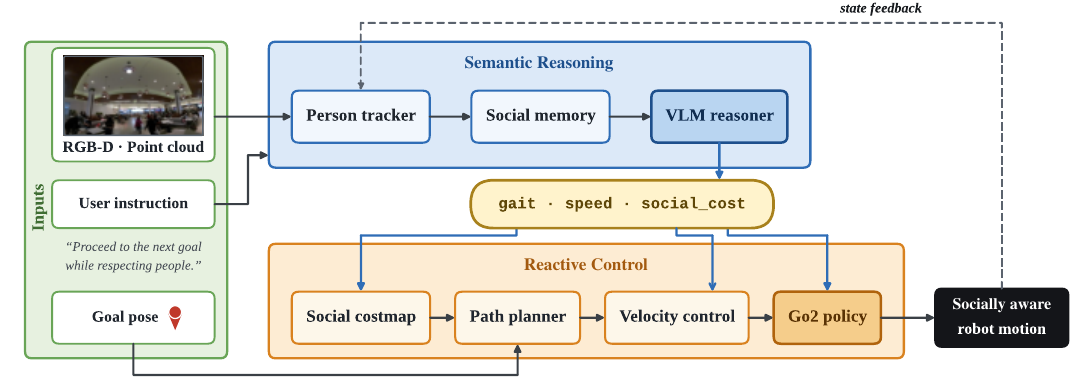}
\vspace{-17.5pt}
\caption{Overview of TACET. The semantic reasoner (person tracker $\rightarrow$ social
memory $\rightarrow$ VLM reasoner) emits the behavior token \token; the reactive controller
(social costmap $\rightarrow$ path planner $\rightarrow$ velocity controller
$\rightarrow$ locomotion policy) consumes it to produce socially aware,
acoustically controlled motion.}
\label{fig:arch}
\vspace{-0.4cm}
\end{figure*}

\subsection{Lightweight VLM Reasoning for Navigation}
Because large VLMs are computationally heavy and slow, a growing body of work
distills their social reasoning into deployable, lightweight models. Social-LLaVA fine-tunes a LLaVA-family VLM on a human-annotated social-reasoning VQA
dataset~\cite{socialllava}; OLiVia-Nav distills context captions into a lightweight model
whose embeddings select socially compliant trajectories~\cite{olivianav}. These efforts build on egocentric and VQA resources
such as SCAND, MuSoHu, JRDB-Social, and SocialNav-SUB~\cite{scand,musohu,jrdbsocial,socialnavsub}.
Our reasoner shares the goal of making a large model's social reasoning cheap enough to run
in the loop, but differs in its output interface: where these produce language, continuous
embeddings, or latent reasoning, we fine-tune a Qwen2.5-VL model to emit a discrete control-
and costmap-conditioning token, an explicit bridge from social reasoning to both low-level
locomotion and the costmap. That token conditions a social costmap and a controllable
acoustic footprint on the same quadruped, without any prior scan of the
environment.

\section{Methodology}
\subsection{Overview}
We consider a quadruped operating in a human-shared indoor space, tasked with
reaching a goal pose while behaving in a context-appropriate manner: respecting the
spatial social norms of nearby people (personal space, conversations, and groups) and
limiting the acoustic intrusion its locomotion imposes on them. These two
objectives are coupled, since how the robot should move depends on who is nearby and
what they are doing, yet they act on different quantities, one spatial and one
acoustic. Rather than collapsing them into a single closed-form cost, TACET infers a
compact behavior token from the scene and uses it to drive a spatial mechanism and
an acoustic mechanism in parallel.

TACET is organized as two processes coupled only through this token (Fig.~\ref{fig:arch}). The semantic reasoner emits $\tau=$~\token, with \texttt{gait}~$\in\{$quiet, normal, urgent$\}$, \texttt{speed}~$\in\{$low, normal, high$\}$, and \texttt{social\_cost} a scalar field amplitude at a resolution of 0.5; the reactive controller consumes it. Neither crosses into the other: the reasoner never outputs motor commands, and the controller never performs semantic reasoning.

The two processes run at different rates so slow reasoning never enters the safety loop. The reasoner emits fresh tokens at roughly 0.25\,Hz and caches them, while the reactive controller---the social costmap, a global path planner, a pure-pursuit velocity controller, and the 50\,Hz locomotion policy---runs continuously. Onboard sensing provides RGB-D, point clouds, and LiDAR-inertial odometry.

Each control cycle (Fig.~\ref{fig:arch}) tracks people into a social memory that retains those out of view (Sec.~\ref{sec:mem}); the memory, the egocentric image, and the instruction go to the VLM (Sec.~\ref{sec:reasoner}), whose token sets the costmap (Sec.~\ref{sec:costmap}) and the gait (Sec.~\ref{sec:loco}). Because the controller always acts on the most recent cached token, a delayed or dropped query degrades the robot to its previous social context rather than an unconditioned one, and the hard obstacle layer is unaffected either way.

\subsection{Semantic Reasoner}
\label{sec:reasoner}
The semantic reasoner maps an egocentric RGB image, a natural-language instruction, and the
social memory to a short chain-of-thought followed by the behavior token $\tau$. Producing $\tau$ as two categorical fields and one scalar amplitude makes the reasoner's output directly parseable and consumable by the
controller, in contrast to free-form language or continuous embeddings.

\textit{Auto-labeling and training data.} Supervision comes from automatically generated labels. A large VLM is prompted to produce structured $\langle$thought, action token$\rangle$ labels on 3{,}950 egocentric frames drawn from MuSoHu~\cite{musohu}, JRDB-Social~\cite{jrdbsocial}, SCAND~\cite{scand}, and InCrowd-VI~\cite{incrowdvi}; to emphasize these robot- and human-perspective social scenes, the auto-labeled records are repeated four times during training. They are combined with 18{,}951 further records: 18{,}042 general images from COCO~\cite{coco} auto-labeled into the same record format, and 909 SocialNav-SUB frames~\cite{socialnavsub} whose existing labels were mapped onto the token format, giving 34{,}751 training records drawn from 22{,}901 distinct source frames. Auto-labeling thus converts heterogeneous scenes into a single, consistent token format. Because the labeling VLM can hallucinate, every generated label is reviewed by a human annotator, who corrects samples whose reasoning or emitted token is inconsistent with the scene, so only verified labels enter training.

\textit{Supervised fine-tuning.} We fine-tune a lightweight open VLM on this mixture with parameter-efficient supervised fine-tuning: the vision encoder and language backbone are frozen and only low-rank adapters are trained, and the loss is applied to the assistant output tokens only. Training on a mixture that up-weights the egocentric sources preserves the base model's general perception while learning the structured navigation interface. We evaluate on a held-out benchmark of 100 SCAND frames taken from sequences excluded from training and labeled by a human annotator, so the reference is independent of the VLM used for labeling. The same VLM without fine-tuning, given the same prompt, matches the reference on 30\% (\texttt{gait}), 26\% (\texttt{speed}), and 14\% (discretized \texttt{social\_cost}), with all three fields agreeing on 14\%. After fine-tuning these rise to 95\%, 96\%, and 69\%, with all three fields agreeing on 65\% of frames. Exact agreement understates the quality of the discretized \texttt{social\_cost}: its mean absolute error decreases from 0.879 to 0.185, and the within-one-bin ($\leq$0.5) agreement increases from 47\% to 96\%.

\textit{Inference.} A query to the reasoner takes 2.6\,s at the median (95th percentile 3.5\,s) on the desktop GPU of Sec.~\ref{sec:impl}, so tokens are refreshed at roughly 0.25\,Hz and cached between queries, and the high-rate controller is never blocked by VLM latency. 

\subsection{Out-of-View Social Memory}
\label{sec:mem}
A narrow camera field of view (FOV) means that people the robot was just reacting
to often leave the frame while remaining well inside its local planning
neighborhood. If the reasoner saw only the current image, an
empty frame would be indistinguishable from ``no one nearby,'' and the behavior
would collapse to a neutral gait even though a person is still close. We address
this with a structured, per-person social memory that persists nearby people across
FOV loss and supplies them to the reasoner as context.

\textit{Tracking and retention.} People are detected with a pose-aware human detector and localized by fusing depth with the 3D point cloud; the resulting tracks are published to the tracker--reasoner bridge, which maintains them in the odometry frame under a hold/coast/refresh policy. A confirmed track is retained while it stays within a $\pm$5.5\,m window around the robot and its observation is recent. While observations continue, including point-cloud-only updates after the detection box disappears, position, velocity, and speed are refreshed from measurement, while orientation is updated conservatively to avoid per-frame yaw noise. Once observations stop, a moving person is coasted forward under a constant-velocity assumption and a stationary person is held in place until a time-to-live expires (12 and 14\,s respectively) or the person leaves the window. A new detection within about 1.6\,m of a coasting track is re-associated to it rather than spawned as a new person, so identity survives brief occlusions.

\textit{Memory record.} Each retained, currently non-visible person is summarized as a structured record: a stable identity, distance and bearing relative to the robot, the time since its last observation, motion state and speed, a social context label (e.g., conversing, grouped, passing), the per-person social cost frozen at the last visible observation, the observation source, and a confidence value. Confidence starts from source quality ($0.88$ visual, $0.76$ point-cloud-only, $0.62$ coasted) and decays exponentially with observation age (time constant ${\approx}18$\,s), which a new live observation resets. The memory therefore carries not just where a person is but how current that estimate is.

\textit{Injection into the reasoner.} The active records are serialized into the VLM prompt alongside the few most recent reasoning outputs, with an explicit instruction to use the current image for visible people and to consult the memory only to decide whether an empty-looking scene is truly clear or merely occluded by the limited FOV. The memory is provided as reasoning context, never as a rule-based override: records for visible people are withheld so an empty image cannot rewrite an occluded person's frozen cost, and the token is always emitted by the reasoner. Compared with a purely geometric buffer feeding a
reactive collision-avoidance layer, our record is socially structured, consumed by the reasoner,
and tags each track with an age-decayed confidence that distinguishes a recent measurement
from an aging prediction.

\subsection{Social Costmap}
\label{sec:costmap}
The social layer represents each person as an anisotropic Gaussian field aligned to the
person's heading, with a larger extent to the front for people in motion, written into one
layer of a layered costmap and scaled by an amplitude and a variance. We use the standard
proxemic-costmap formulation for this field~\cite{kirby,layeredcostmap}; what TACET changes
is where those two parameters come from. Prior work sets them offline, tying them to a hand-specified cue such as the
person's walking speed~\cite{kirby} or a discrete activity label~\cite{adaptiveproxemic}.
In TACET they follow the emitted token, so the same decision that sets how quietly the
robot walks also sets how much room it leaves. Collisions remain in a separate hard
obstacle layer, so social cost modulates preference without compromising geometric
safety.

\textit{Per-person field.} Each person induces an anisotropic 2-D Gaussian in their
own frame,
\begin{equation}
\label{eq:field}
c(\mathbf{p}) = A\,\exp\!\Big(-\tfrac{1}{2}\big((u+\delta)^2/\sigma_{\mathrm{long}}^2 + v^2/\sigma_{\mathrm{side}}^2\big)\Big),
\end{equation}
with $\sigma_{\mathrm{long}}=\sigma_{\mathrm{front}}$ for $u\ge0$ and
$\sigma_{\mathrm{rear}}$ for $u<0$, where $u$ and $v$ are the longitudinal and
lateral offsets in the person frame and $\delta$ shifts the field center slightly
behind the person. Splitting $\sigma_{\mathrm{long}}$ into a larger front and smaller rear extent yields the
familiar egg-shaped personal space~\cite{kirby} oriented along the person's travel
direction for a person in motion, and a circular field otherwise.

\textit{Token $\rightarrow$ amplitude.} The \texttt{social\_cost} token sets the
field strength. Writing $s$ for the token value after a fixed gain and
exponential-moving-average filtering, we map
\begin{equation}
\label{eq:amp}
\begin{aligned}
\rho &= \max\!\big(\rho_{\mathrm{floor}},\,\min(1,\,s/s_{\max})\big),\\
A   &= A_0 + (A_{\max}-A_0)\,\rho,
\end{aligned}
\end{equation}
so that more cautious situations (larger $s$) raise the amplitude toward
$A_{\max}$, while a floor $\rho_{\mathrm{floor}}$ guarantees a minimum clearance
even when $s$ is small.

\textit{Token $\rightarrow$ width and motion.} The same normalized $\rho$, together
with gait and speed multipliers $M_{\mathrm{gait}}, M_{\mathrm{speed}}$, and a
per-person scale $k_{\mathrm{shape}}$, sets the field width, and a motion signal
$m\in[0,1]$ elongates the front lobe for people in motion, above a small
threshold $m_0$:
\begin{equation}
\label{eq:width}
\begin{gathered}
\sigma = \sigma_0(1+0.58\,\rho)\,M_{\mathrm{gait}}M_{\mathrm{speed}}\,k_{\mathrm{shape}},\\[2pt]
\sigma_{\mathrm{side}} = \sigma_{\mathrm{rear}} = \sigma,\\[2pt]
\sigma_{\mathrm{front}} = \sigma\,\big(1+0.45\,m\big)\ \text{if } m>m_0,\ \text{else } \sigma.
\end{gathered}
\end{equation}
Using the observed motion direction rather than an estimated body yaw avoids the
brittleness of per-frame pose estimation while still extending cost ahead of moving
people. The coefficients in \eqref{eq:width} were tuned once in simulation and
never re-tuned per scenario.

\textit{Group field.} Following group-aware proxemic models, which represent an interacting
group as one zone rather than a sum of individual ones~\cite{groupgaussian}, we merge
people judged to be interacting (pairwise spacing within a small range) into a single
field spanning the segment connecting them, protecting the shared interaction space
(o-space):
\begin{equation}
\label{eq:group}
c_{\mathrm{grp}} = A_{\mathrm{grp}}\exp\!\Big(-\tfrac12\big(d_\perp^2/\sigma_{\mathrm{lat}}^2 + d_\parallel^2/\sigma_{\mathrm{end}}^2\big)\Big),
\end{equation}
with $A_{\mathrm{grp}}=\min(A_{\max},1.25\,A)$, where $d_\perp$ is the perpendicular offset
from the connecting segment, $d_\parallel$ the distance beyond either endpoint (zero
between them), $\sigma_{\mathrm{lat}}$ a fixed lateral extent (0.65\,m), and
$\sigma_{\mathrm{end}}=\max(0.35\,\mathrm{m},\,0.35\,d)$ set by the inter-person spacing
$d$. The amplitude again follows the inferred token, so the same pair induces different
avoidance depending on the social context.

\subsection{Quiet Locomotion}
\label{sec:loco}
The \texttt{gait} and \texttt{speed} tokens select how the robot moves. Because
quadruped locomotion noise is dominated by foot--ground impacts, whose acoustic
energy grows with foot velocity at contact~\cite{mute}, we reduce perceived noise
through control rather than hardware by shaping the policy's foot-contact behavior
under token conditioning.

\textit{Policy.} A single policy maps a 64-dimensional observation (proprioception, the velocity command, and the token descriptor with its gait clock, excluding base linear velocity) through an MLP with hidden sizes $[512,256,128]$ to twelve joint position targets at 50\,Hz, tracked by joint PD control. It is trained with reinforcement learning in massively parallel simulation and deployed zero-shot on the robot. The gait and speed tokens enter the policy as physical quantities rather than a one-hot
code: the gait token selects a contact schedule (a four-beat crawl for quiet; a trot with
decreasing duty factor for normal and urgent) and the speed token its period, so one network
covers every gait--speed pair the reasoner emits (quiet at low or normal speed, normal at
any speed, urgent at high speed).

\textit{Reward.} The reward combines velocity tracking with token-dependent style and acoustic terms and standard stability/smoothness regularizers. The key acoustic terms are a token-gated foot-impact penalty that tightens as the gait token moves from urgent to quiet at matched speed, together with contact-force and loading-rate penalties that cap the loudest footfalls; gait-style terms (contact-schedule tracking, clearance, and air time) keep each commanded gait well-formed; and a left--right symmetry objective, an LR joint-mirror penalty together with a yaw-rate penalty on straight-line commands, suppresses lateral drift. Unlike an operator-set quiet factor~\cite{mute}, the gait is chosen by a socially inferred token, so quietness is allocated by context rather than preset.

\section{Experiments}
We evaluate TACET at two levels. We first characterize the locomotion policy as a noise
source in isolation, on a controlled traverse with no people present, separating the
gait's acoustic contribution from that of any detour. We then evaluate
the full method in social encounters, measuring whether the
robot is simultaneously space-respecting and quiet toward the people it meets. Each
condition is repeated for ten trials and we report the mean, following the
evaluation protocol of prior quiet-locomotion work~\cite{mute}.

\subsection{Implementation Details}
\label{sec:impl}
The platform is a Unitree Go2 with an Intel RealSense D435i RGB-D camera and a Hesai XT-16
LiDAR; VLM inference runs on a single NVIDIA RTX A6000 (48\,GB). YOLOv8n-pose~\cite{yolov8},
Point-LIO~\cite{pointlio}, and Nav2's Smac Hybrid-A*~\cite{nav2} run at default settings. Labels are generated with Gemini~2.5
Flash~\cite{gemini}; the reasoner is Qwen2.5-VL-7B~\cite{qwenvl} fine-tuned with 4-bit
QLoRA~\cite{qlora} for one epoch. The policy is trained with PPO in Isaac Lab~\cite{isaaclab}
(${\approx}25{,}000$ iterations, 4{,}096 parallel environments) and deployed through
rl\_sar~\cite{rlsar}. Further implementation details, costmap and reward parameters, and
the human-study questionnaire and protocol are on the project page.

\begin{table}[t]
\captoppad
\caption{Locomotion noise at matched speeds.}
\label{tab:axisl}
\centering\small
\setlength{\tabcolsep}{1pt}
\begin{threeparttable}
\begin{tabular}{>{\raggedright\arraybackslash}p{1.90cm}>{\raggedright\arraybackslash}p{2.30cm}>{\centering\arraybackslash}p{2.06cm}>{\centering\arraybackslash}p{2.06cm}}
\toprule
Speed (m/s) & Method & MNL (dBA)$\,\downarrow$ & PNL (dBA)$\,\downarrow$ \\
\midrule
\multirow{3}{*}{0.5}
 & Default Go2 & $68.9 \pm 2.8$ & $72.5 \pm 0.7$ \\
 & MUTE ($\beta{=}1$) & $67.8 \pm 2.0$ & $71.5 \pm 0.5$ \\
 & \cellcolor{oursrow}\textbf{Ours} & \cellcolor{oursrow}$\mathbf{64.1 \pm 1.3}$ & \cellcolor{oursrow}$\mathbf{65.9 \pm 0.8}$ \\
\midrule
\multirow{3}{*}{0.8}
 & Default Go2 & $74.4 \pm 2.4$ & $77.2 \pm 1.2$ \\
 & MUTE ($\beta{=}1$) & $71.7 \pm 2.9$ & $74.6 \pm 0.7$ \\
 & \cellcolor{oursrow}\textbf{Ours} & \cellcolor{oursrow}$\mathbf{65.1 \pm 2.1}$ & \cellcolor{oursrow}$\mathbf{68.1 \pm 1.6}$ \\
\midrule
\multirow{3}{*}{1.1}
 & Default Go2 & $74.8 \pm 1.9$ & $76.9 \pm 1.2$ \\
 & MUTE ($\beta{=}1$) & $72.3 \pm 2.5$ & $76.6 \pm 0.2$ \\
 & \cellcolor{oursrow}\textbf{Ours} & \cellcolor{oursrow}$\mathbf{70.6 \pm 2.4}$ & \cellcolor{oursrow}$\mathbf{75.1 \pm 0.6}$ \\
\bottomrule
\end{tabular}
\begin{tablenotes}[flushleft]\scriptsize
\item[] Speed is the commanded speed in m/s; MNL and PNL are in dBA (lower is
better). Values are mean $\pm$ SD over ten trials; bold indicates best.
\vspace{-0.4cm}
\end{tablenotes}
\end{threeparttable}
\end{table}

\subsection{Acoustic Locomotion Benchmark}
\textit{Setup.} Following the quiet-locomotion evaluation protocol~\cite{mute}, the
robot traverses at three fixed commanded speeds (0.5, 0.8, and 1.1\,m/s) over a path
of at least 10\,m, with ten trials per speed; we report per-speed results.
Because here we characterize the policy
as a noise \emph{source} rather than a listener's exposure, a calibrated
sound-level meter is mounted on top of the robot at a fixed offset
(0.05\,m above the body) with no people present. Baselines are compared at
\emph{matched actual} speed, taken from the LiDAR-inertial odometry~\cite{pointlio}
rather than from leg odometry, whose contact assumption makes its drift
gait-dependent; a method therefore cannot appear quieter merely by moving more
slowly. Standing still with only its cooling fan running, the Go2 already registers
about 60\,dBA at the same microphone position, the floor for the
locomotion levels in Table~\ref{tab:axisl}.

\textit{Why A-weighting.} A-weighting attenuates the low frequencies to which hearing is
least sensitive, approximating \emph{perceived} loudness rather than total radiated
energy~\cite{iec61672}, as is conventional for quiet legged locomotion~\cite{mute}. We report the Mean and Peak Noise Level (MNL, PNL), the time-averaged and maximum level over a traverse.

\textit{Baselines.} We compare the default Go2 controller, our reimplementation of MUTE~\cite{mute} at $\beta{=}1$, its quietest setting, and ours, which emits the gait token appropriate to each commanded speed (Table~\ref{tab:axisl}).

Across all three speeds, Ours is the quietest at matched speed, lowering MNL by up to 9.3\,dBA over the default controller and 6.6\,dBA over MUTE (at 0.8\,m/s), with a similar ordering in PNL. As expected, absolute noise rises with speed for every method, since faster foot--ground impacts are louder, but Ours stays below both baselines at each speed, showing that the per-speed gait token yields a consistent quieting margin rather than a single fixed setting. The margin is smallest at 1.1\,m/s, where impact noise dominates and little headroom remains for quieting.

\subsection{System-Level Social Navigation}
\begin{table}[t]
\captoppad
\caption{Social-navigation results across four scenarios.}
\label{tab:axiss}
\centering\footnotesize
\setlength{\tabcolsep}{1pt}
\begin{threeparttable}
\begin{tabular}{>{\raggedright\arraybackslash}p{1.55cm}>{\raggedright\arraybackslash}p{1.28cm}>{\centering\arraybackslash}p{1.10cm}>{\centering\arraybackslash}p{1.62cm}>{\centering\arraybackslash}p{1.42cm}>{\centering\arraybackslash}p{1.10cm}}
\toprule
Scenario & Method & PSC$\,\uparrow$ & MinDist$\,\uparrow$ & $\Delta L\,\downarrow$ & AID$\,\downarrow$ \\
\midrule
\multirow{5}{*}{\shortstack[l]{Focused\\(working)}}
 & RPP            & 95.6 & $1.03\pm0.12$ & $8.4\pm0.5$ & 66.3 \\
 & SFM            & 92.7 & $0.41\pm0.05$ & $5.5\pm0.3$ & 51.8 \\
 & MUTE           & 97.4 & $0.91\pm0.14$ & $6.4\pm0.3$ & 61.2 \\
 & VLM-SN         & \textbf{100.0} & $1.35\pm0.06$ & $5.9\pm0.2$ & 58.3 \\
 & \cellcolor{oursrow}\textbf{Ours} & \cellcolor{oursrow}\textbf{100.0} & \cellcolor{oursrow}$\mathbf{3.61\pm0.17}$ & \cellcolor{oursrow}$\mathbf{1.6\pm0.2}$ & \cellcolor{oursrow}\textbf{0.0} \\
\midrule
\multirow{5}{*}{\shortstack[l]{Focused\\(headph.)}}
 & RPP            & 95.8 & $1.06\pm0.14$ & $8.1\pm0.6$ & 65.9 \\
 & SFM            & 92.2 & $0.39\pm0.06$ & $5.7\pm0.2$ & 52.3 \\
 & MUTE           & 97.1 & $0.89\pm0.15$ & $6.3\pm0.3$ & 60.1 \\
 & VLM-SN         & \textbf{100.0} & $1.35\pm0.04$ & $5.7\pm0.2$ & 57.9 \\
 & \cellcolor{oursrow}\textbf{Ours} & \cellcolor{oursrow}\textbf{100.0} & \cellcolor{oursrow}$\mathbf{3.62\pm0.19}$ & \cellcolor{oursrow}$\mathbf{2.9\pm0.2}$ & \cellcolor{oursrow}\textbf{26.5} \\
\midrule
\multirow{5}{*}{\shortstack[l]{Standing\\group}}
 & RPP            & 92.3 & $0.90\pm0.19$ & $12.1\pm1.1$ & 100.0 \\
 & SFM            & 92.8 & $1.02\pm0.02$ & $7.8\pm0.6$ & 73.3 \\
 & MUTE           & 93.7 & $1.13\pm0.05$ & $4.9\pm0.5$ & 42.9 \\
 & VLM-SN         & 90.4 & $0.87\pm0.16$ & $7.3\pm0.5$ & 65.7 \\
 & \cellcolor{oursrow}\textbf{Ours} & \cellcolor{oursrow}\textbf{100.0} & \cellcolor{oursrow}$\mathbf{2.25\pm0.21}$ & \cellcolor{oursrow}$\mathbf{2.7\pm0.3}$ & \cellcolor{oursrow}\textbf{22.3} \\
\midrule
\multirow{5}{*}{\shortstack[l]{Moving\\person}}
 & RPP            & 87.7 & $0.59\pm0.05$ & $7.9\pm0.7$ & 83.2 \\
 & SFM            & 90.2 & $0.75\pm0.04$ & $5.1\pm0.5$ & 51.3 \\
 & MUTE           & 98.9 & $1.21\pm0.11$ & $3.8\pm0.3$ & 24.7 \\
 & VLM-SN         & 94.1 & $0.69\pm0.07$ & $4.1\pm0.4$ & 29.4 \\
 & \cellcolor{oursrow}\textbf{Ours} & \cellcolor{oursrow}\textbf{100.0} & \cellcolor{oursrow}$\mathbf{1.44\pm0.12}$ & \cellcolor{oursrow}$\mathbf{1.7\pm0.3}$ & \cellcolor{oursrow}\textbf{0.0} \\\bottomrule
\end{tabular}
\begin{tablenotes}[flushleft]\scriptsize
\item[] PSC in \%, MinDist in m, $\Delta L$ (sound-level increment) in dBA, AID in \%;
VLM-SN: VLM-Social-Nav~\cite{vlmsocialnav}. Means over ten trials; MinDist and $\Delta L$ are
mean $\pm$ SD. Bold indicates best.
\vspace{-0.15cm}
\end{tablenotes}
\end{threeparttable}
\end{table}

\textit{Metrics.} We report four objective metrics spanning the spatial-social and
acoustic-social dimensions:
\begin{itemize}
\item \textbf{Personal Space Compliance (PSC, \%, $\uparrow$).} Fraction of the run for which the robot keeps a center-to-center distance of at least 1.2\,m from every person, which is Hall's \emph{personal}/\emph{social} boundary~\cite{hall} and a standard proxemic proxy~\cite{francis,falcon}.
\item \textbf{Minimum Distance (MinDist, m, $\uparrow$).} The smallest center-to-center robot--person distance reached during the run. PSC measures
how \emph{long} the robot stays outside the personal zone, MinDist how \emph{deeply} it
intrudes at the closest point; a run can score well on one and poorly on the other, so we
report both~\cite{francis}.
\item \textbf{Sound-Level Increment ($\Delta L$, dBA, $\downarrow$).} With
$L_{\mathrm{on}}$ the A-weighted level at the listener while the robot operates and
$L_{\mathrm{amb}}$ the matched ambient level, we report
$\Delta L=L_{\mathrm{on}}-L_{\mathrm{amb}}$. It is the increase in total listener level due
to robot operation, not an estimate of robot-only SPL. The meter sits at the person's location;
$L_{\mathrm{amb}}$ is 49\,dBA in the two focused-person conditions, 47\,dBA for the group,
and 53\,dBA outdoors.
\item \textbf{Acoustic Intrusion Duration (AID, \%, $\downarrow$).} Fraction of the run for which $\Delta L$ exceeds 5\,dBA, the level at which a change is
clearly noticeable rather than barely perceptible~\cite{perception}, so AID captures how
long the robot is \emph{audibly} above ambient.
\end{itemize}

\textit{Scenarios.} Seven encounters test context-dependent behavior. The first four place
people on the route and use the social metrics above; the last three do not.
\begin{itemize}
\item \textbf{Focused person, working (indoor).} A stationary worker in a quiet indoor space tests whether the robot avoids acoustic intrusion.
\item \textbf{Focused person, headphones (indoor).} The same scene with only headphones added isolates a cue that relaxes acoustic, but not spatial, caution.
\item \textbf{Standing group (indoor).} A conversing group blocks the route, requiring the robot to avoid its shared space.
\item \textbf{Moving person (outdoor).} A crossing pedestrian tests yielding under higher ambient noise.
\item \textbf{Empty corridor (indoor).} With no person present, any detour or slowdown is overhead.
\item \textbf{Open sports field (outdoor).} Distant activity leaves the robot's route unobstructed.
\item \textbf{Emergency (indoor).} A flashing red siren in an empty, dim corridor tests whether urgency overrides acoustic restraint.
\end{itemize}

\textit{Baselines.} All baselines run on the same Go2 and planning stack. \textbf{RPP:} the Nav2 Regulated Pure Pursuit controller treats people as ordinary obstacles with the default gait. \textbf{SFM:} a classical social planner whose proxemic repulsion shapes the path~\cite{sfm}, default gait. \textbf{MUTE:} the quiet locomotion policy driven by the RPP controller, with no social reasoning~\cite{mute}. \textbf{VLM-Social-Nav:} a VLM-based social planner that scores candidate actions with a vision-language model~\cite{vlmsocialnav}, using GPT-4o as the reasoning model, default gait; it reasons about social context but not about the robot's own acoustic footprint. \textbf{Ours} is the complete TACET method.

\begin{table}[!t]
\captoppad
\caption{Time to goal and emitted tokens by scenario.}
\label{tab:time}
\centering\small
\setlength{\tabcolsep}{3pt}
\begin{threeparttable}
\begin{tabular*}{\linewidth}{@{\extracolsep{\fill}}lccc|ccc@{}}
\toprule
\multirow{2}{*}{Scenario} & \multicolumn{3}{c|}{Time to goal (\textnormal{s})} & \multicolumn{3}{c}{Emitted token} \\
\cmidrule(lr){2-4}\cmidrule(l){5-7}
 & mean & best & \textbf{Ours} & \texttt{gait} & \texttt{spd} & \texttt{cost} \\
\midrule
Focused (working) & 31.8 & \textbf{21.2} & \cellcolor{oursrow}36.8 & quiet & low & 4.0 \\
Focused (headph.) & 31.8 & \textbf{22.0} & \cellcolor{oursrow}33.8 & normal & low & 4.0 \\
Standing group & 17.2 & \textbf{13.6} & \cellcolor{oursrow}25.6 & quiet & low & 3.5 \\
Moving person & 16.7 & \textbf{14.8} & \cellcolor{oursrow}21.8 & quiet & normal & 2.5 \\
\midrule
Empty corridor & 16.2 & \textbf{14.4} & \cellcolor{oursrow}\textbf{14.4} & normal & normal & 1.5 \\
Sports field & 18.3 & 14.4 & \cellcolor{oursrow}\textbf{14.0} & normal & high & 1.0 \\
Emergency & 15.4 & 14.2 & \cellcolor{oursrow}\textbf{11.2} & urgent & high & 1.0 \\
\bottomrule
\end{tabular*}
\begin{tablenotes}[flushleft]\scriptsize
\item[]\raggedright Times are trial means; tokens are trial modes. ``Baseline mean'' averages
RPP, SFM, MUTE, and VLM-Social-Nav, whereas ``Baseline best'' is the fastest of
those four; Ours is excluded from both baseline summaries. Bold marks the fastest
entry in each row. The midrule separates person-facing and no-person scenarios.
\vspace{-0.4cm}
\end{tablenotes}
\end{threeparttable}
\end{table}

Table~\ref{tab:axiss} reports the per-scenario results. Ours is the only method that keeps PSC at 100\% in all four
scenarios; VLM-Social-Nav reaches 100\% with the focused person but not with the group or the moving person.
Ours also keeps a far wider margin at the closest approach, with MinDist never below
1.44\,m against at most 1.35\,m for any baseline.

All runs reached the goal without contacting a person, but not always because the robot
made room. SFM oscillated in the narrow passage around the focused person, scraping the
wall and nearly brushing the person; RPP did not yield to the moving person and stayed
clear only because the person stepped aside. Three baseline conditions came within arm's reach of
a person: SFM in both focused-person conditions (MinDist 0.41 and 0.39\,m) and RPP with the
moving person (0.59\,m). Ours is also the quietest in every scenario and audible above ambient for the smallest
fraction of the run.

\textit{What the reasoner adds.} The two focused-person conditions hold the scene fixed and
change only the headphones. Methods without token-conditioned gait control behave
near-identically in both; VLM-Social-Nav, for example, yields $\Delta L$ 5.9 vs.\ 5.7\,dBA.
Ours separates them while maintaining essentially the same clearance
(MinDist 3.61 vs.\ 3.62\,m) and relaxing the acoustic effort where the listener is less
sensitive to it ($\Delta L$ 1.6 vs.\ 2.9\,dBA, AID 0.0 vs.\ 26.5\%). With the closest
approach unchanged and only the gait token differing, the acoustic difference is
consistent with the gait effect measured in Table~\ref{tab:axisl} (Sec.~\ref{sec:abl}).

Table~\ref{tab:time} shows that the cost of this behavior varies with what the context
calls for: where people need space or quiet, TACET spends at most 1.5 times the mean
baseline time, but in an empty corridor it matches the fastest baseline (14.4\,s) and on an
open sports field it is the fastest method (14.0\,s against 14.4\,s), as it is with
the siren light in the corridor (11.2\,s against 14.2\,s). The token columns show why: the instruction names neither gait, speed, nor clearance, so the token follows the scene and the behavior follows the
token.

\subsection{Human Study}
Following prior social-navigation user studies~\cite{raj,vlmsocialnav}, we ran a blinded
video study with $N{=}15$ observers. Each observer
rated the 20 clips formed by four person-facing scenarios and five methods, presented in
randomized order. After each
clip, observers rated Q1--Q4 on a 0--5 Likert scale.

\begin{figure}[t]
\centering
\includegraphics[width=\columnwidth]{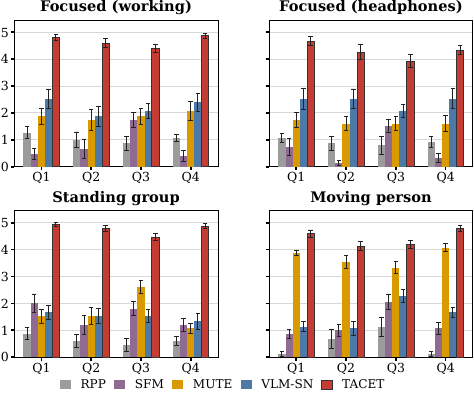}
\vspace{-0.53cm}
\caption{Per-question results of the blinded video study. Q1: safe and comfortable
distance; Q2: contextual consideration; Q3: footstep-noise appropriateness; Q4:
scenario-specific etiquette. Bars show mean scores on a 0--5 Likert scale; error bars
show SEs. Higher is better.}
\label{fig:humanstudy}
\vspace{-0.15cm}
\end{figure}

Fig.~\ref{fig:humanstudy} plots the per-question averages, taken over paired
within-observer ratings, with standard errors for the five methods in each scenario. TACET
receives the highest agreement in all sixteen scenario--question cells, averaging 4.54
against 2.23 for MUTE, 1.92 for VLM-Social-Nav, 1.07 for SFM and 0.78 for RPP, and its
standard errors are the smallest (0.14 on average against 0.29 for VLM-Social-Nav), indicating
more consistent ratings across observers. The margin narrows only with the
moving person, where MUTE reaches 3.33--4.07; that is also where MUTE is the strongest
baseline on the objective metrics (Table~\ref{tab:axiss}), consistent with its observer ratings there.

\subsection{Ablation Study}
\label{sec:abl}
\begin{table}[!t]
\captoppad
\caption{Ablations on two person-facing scenarios.}
\label{tab:ablation}
\centering\footnotesize
\setlength{\tabcolsep}{1pt}
\begin{threeparttable}
\begin{tabular}{>{\raggedright\arraybackslash}p{1.16cm}>{\raggedright\arraybackslash}p{2.00cm}>{\centering\arraybackslash}p{1.00cm}>{\centering\arraybackslash}p{1.56cm}>{\centering\arraybackslash}p{1.36cm}>{\centering\arraybackslash}p{1.00cm}}
\toprule
Scenario & Variant & PSC$\,\uparrow$ & MinDist$\,\uparrow$ & $\Delta L\,\downarrow$ & AID$\,\downarrow$ \\
\midrule
\multirow{6}{*}{\shortstack[l]{Focused\\(working)}}
 & Qwen2.5-VL-7B & \textbf{100.0} & $1.35\pm0.10$ & $4.5\pm0.5$ & 31.8 \\
 & InternVL3.5-8B & \textbf{100.0} & $1.33\pm0.12$ & $4.9\pm0.4$ & 35.1 \\
\cmidrule(lr){2-6}
 & \cellcolor{oursrow}\textbf{Ours (full)} & \cellcolor{oursrow}\textbf{100.0} & \cellcolor{oursrow}$\mathbf{3.61\pm0.17}$ & \cellcolor{oursrow}$\mathbf{1.6\pm0.2}$ & \cellcolor{oursrow}\textbf{0.0} \\
 & w/o soc. layer & 96.5 & $1.18\pm0.11$ & $4.3\pm0.3$ & 29.3 \\
 & w/o trained pol. & \textbf{100.0} & $3.47\pm0.19$ & $7.4\pm0.5$ & 63.3 \\
 & w/o soc. memory & \textbf{100.0} & $3.54\pm0.24$ & $3.9\pm0.3$ & 18.7 \\
\midrule
\multirow{6}{*}{\shortstack[l]{Moving\\person}}
 & Qwen2.5-VL-7B & \textbf{100.0} & $1.38\pm0.11$ & $3.5\pm0.5$ & 36.1 \\
 & InternVL3.5-8B & \textbf{100.0} & $1.33\pm0.12$ & $3.8\pm0.7$ & 39.3 \\
\cmidrule(lr){2-6}
 & \cellcolor{oursrow}\textbf{Ours (full)} & \cellcolor{oursrow}\textbf{100.0} & \cellcolor{oursrow}$\mathbf{1.44\pm0.12}$ & \cellcolor{oursrow}$\mathbf{1.7\pm0.3}$ & \cellcolor{oursrow}\textbf{0.0} \\
 & w/o soc. layer & 94.7 & $0.89\pm0.12$ & $3.8\pm0.3$ & 37.4 \\
 & w/o trained pol. & \textbf{100.0} & $1.39\pm0.13$ & $4.2\pm0.5$ & 41.5 \\
 & w/o soc. memory & \textbf{100.0} & $1.42\pm0.09$ & $2.6\pm0.4$ & 19.3 \\
\bottomrule
\end{tabular}
\begin{tablenotes}[flushleft]\scriptsize
\item[] PSC and AID in \%, MinDist in m, $\Delta L$ in dBA; means over ten trials; MinDist and
$\Delta L$ are mean $\pm$ SD; bold indicates best. Rows above each rule swap only the reasoner that
emits $\tau$; rows below remove one component.
\vspace{-0.4cm}
\end{tablenotes}
\end{threeparttable}
\end{table}

We remove one component of TACET at a time, keeping the rest fixed, and repeat this on two
representative person-facing scenarios (Table~\ref{tab:ablation}). Removing the social layer disables
\eqref{eq:field}--\eqref{eq:group} while the trained policy is kept, so people are planned
around as ordinary obstacles; removing the trained policy (default gait, with the social layer
and social memory intact) is its mirror image, keeping the socially aware detour but not the
token-conditioned policy. Because people pass in and out of the camera FOV as the robot moves,
the \emph{w/o social memory} variant isolates whether behavior stays consistent across these
brief occlusions. Separately, the top block of each scenario keeps the architecture fixed and
swaps only the reasoner that emits $\tau$, isolating the contribution of fine-tuning beyond
prompt engineering.

Removing the social layer costs both compliance and clearance in each scenario, dropping
MinDist inside the 1.2\,m zone the full method never enters. The acoustic terms degrade
with it: with people treated as ordinary obstacles the robot passes closer, so the same
gait is louder at the listener. Removing the trained policy instead leaves the route
essentially unchanged---both MinDist differences fall within one standard
deviation---while $\Delta L$ and AID degrade the most of any variant, supporting an acoustic
contribution from the gait policy beyond that of the detour. Removing social memory alters
the route only marginally (3.54 and 1.42\,m), yet the robot is audible above ambient for
18.7\% and 19.3\% of the run against 0.0\% for the full method: without the memory a
person who briefly leaves the camera view is no longer carried in the reasoner's context,
so the policy is no longer conditioned on a quiet token across the occlusion.

Swapping the reasoner degrades both axes. With the focused person, a prompted
Qwen2.5-VL-7B~\cite{qwenvl} keeps only 1.35\,m of clearance against 3.61\,m for the full
method, close to what removing the social layer gives, and raises $\Delta L$ to 4.5\,dBA with
AID at 31.8\%; InternVL3.5-8B~\cite{internvl35} behaves the same way (1.33\,m, 4.9\,dBA,
35.1\%). With the moving person the route changes far less (1.38 and 1.33\,m against
1.44\,m), yet the robot is audible above ambient for 36.1\% and 39.3\% of the run against
0.0\% for the full method. Since the architecture is unchanged, the loss follows from the
token interface itself rather than from any component, consistent with the base model's low token
agreement in Sec.~\ref{sec:reasoner}.

These variants also separate distance from gait. Because Ours both leaves more room and
walks more quietly, its lower $\Delta L$ in Table~\ref{tab:axiss} might be read as a
by-product of the larger MinDist. The \emph{w/o trained policy} variant keeps the closest approach within one standard
deviation of the full method yet raises $\Delta L$ by 5.8 and 2.5\,dBA. The cleanest isolation is
Table~\ref{tab:axisl}, which removes geometry altogether (the meter rides on the robot and
no one is present), yet gaits still differ by up to 9.3\,dBA at matched speed. $\Delta L$
is integrated over the whole run rather than read at the closest approach, so a shared
MinDist does not by itself fix the acoustic outcome.

\section{Conclusion}
We presented TACET, which treats a quadruped's footstep noise as something the robot
decides rather than a fixed property of its hardware. One inferred token sets both where
it walks and how loudly, improving both spatial and acoustic metrics in the
evaluated scenarios, and the ablations support an acoustic contribution from the gait
policy beyond that of the detour. Quietness is therefore allocated by context rather than preset, with the reasoner
emitting a quiet gait near a focused worker and an urgent one in the emergency scenario.

Limitations include the reasoner's low update rate, which the reactive costmap partially mitigates but which we do not evaluate under rapid scene changes. Socially aware detours also increase time to goal, and we do not evaluate their effect on people sharing the space. Finally, the scenario set is limited. Future work targets faster onboard reasoning and broader evaluation in more dynamic, populated settings.

\bibliographystyle{IEEEtran}
\bibliography{references}

\end{document}